\documentclass{article}
\usepackage{setspace} 

\usepackage{amssymb}
\usepackage[affil-it]{authblk}
\usepackage{amsthm}
\usepackage{breqn}

\usepackage{verbatim}
\usepackage{multirow}
\usepackage{authblk}

\usepackage{soul} 

\begin{document}




\title{Interpretable pap smear cell representation for cervical cancer screening}


\author[1]{Yu Ando}
\author[3,4]{Nora Jee-Young Park}
\author[5,6]{Gun Oh Chong}
\author[1]{Seokhwan Ko}
\author[1]{Donghyeon Lee}
\author[2]{Junghwan Cho  \thanks{Corresponding author: joshua@knu.ac.kr}}
\author[1,2]{Hyungsoo Han}


\affil[1]{Department of Biomedical Science, Kyungpook National University}
\affil[2]{Clinical Omics Institute, Kyungpook National University}
\affil[3]{Department of Pathology, School of Medicine, Kyungpook National University}
\affil[4]{Department of Pathology, Kyungpook National University Chilgok Hospital}
\affil[5]{Department of Obstetrics and Gynecology, School of Medicine, Kyungpook National University}
\affil[6]{Department of Obstetrics and Gynecology, Kyungpook National University Chilgok Hospital}
\date{}

\maketitle
\begin{abstract}
    Screening is critical for prevention and early detection of cervical cancer but it is time-consuming and laborious. Supervised deep convolutional neural networks have been developed to automate pap smear screening and the results are promising. However, the interest in using only normal samples to train deep
    neural networks has increased owing to class imbalance problems and high-labeling costs that are both prevalent in healthcare.
    In this study, we introduce a method to learn explainable deep cervical cell representations for pap smear cytology images based on one class classification using variational autoencoders. 
    
    Findings demonstrate that a score can be calculated for cell abnormality without training models with abnormal samples and localize abnormality to interpret our results with a novel metric based on absolute difference in cross entropy in agglomerative clustering.
    The best model that discriminates squamous cell carcinoma (SCC) from normals gives $0.908\pm 0.003$ area under operating characteristic curve (AUC) and one that
    discriminates high-grade epithelial lesion (HSIL) $0.920\pm 0.002$ AUC. Compared to other clustering methods, our method enhances the V-measure and yields higher homogeneity scores, which more effectively isolate different abnormality regions, aiding in the interpretation of our results. Evaluation using in-house and additional open dataset show that our model can discriminate abnormality without the need of additional training of deep models.

\end{abstract}






\section{Introduction}
\label{sec:sample1}
Cervical cancer screening is an invaluable test for early detection and prevention of cervical cancer \cite{jansen2020effect}. The Pap smear is a screening method that examines cervical cells to discern abnormalities \cite{bedell2020cervical}. While pap smear screening has been successful at reducing rates of cervical cancer mortality, incidence and mortality increased in developing countries owing to the lack of resources \cite{sung2021global}. Developing quantitative and autonomous methods using computer technology has the potential to improve healthcare worldwide by making the smear screening process cost-efficient.

Recently, large models trained using supervised deep learning were proposed and the obtained results have been promising \cite{basak2021cervical,n2021deep,manna2021fuzzy}. Despite the positive outlook, models were limited in providing explanations for their predictions. Additionally, training these models requires many labeled datasets. In the medical domain, it is expensive to acquire vast amounts of labeled data due to the expertise required to interpret and generate data.

One-class classification (OCC) \cite{moya1993one} is another machine-learning problem where the objective is to train a model that can distinguish objects belonging to a specific class, called the normal class, from all other objects or instances that are considered anomalous. OCC has gained significant attention in recent years owing to its potential applications in various domains, including cybersecurity \cite{kim2020anomaly}, fault detection \cite{lee2020fault}, and medical tasks \cite{gao2020handling}. Owing to the regular screening of cervical cancer \cite{lim2019current}, we can expect more samples of normal-cell samples compared with the abnormal ones. This suggests that OCC may be an invaluable method for screening. The goal is to capture the intrinsic characteristics of the normal class and identify abnormalities (deviations from normality).

Notwithstanding promising methods in deep learning, the inherent black-box nature
of deep models prevents the user from knowing how the model made a prediction or inference. Models that can explain their predictions may gain the trust of service providers and users and may support those involved with informed decisions in fields that require informed consent or in cases entailing high-stake decisions. Deep-learning methods that produce reasons for their predictions will be more valuable in these domains.

Disentanglement is the property of representation where a unit of the representation, an entry or entries of a feature vector, correspond to a factor of variation of data \cite{bengio2013representation}. For example, if a model is trained on images of human faces, a disentangled representation may give entries that correspond to age, hair length, facial color, etc. A model that has such property can be used to interpret results by finding which feature is most important. Variational autoencoder (VAE) based models \cite{kingma2013auto,burgess2018understanding,chen2018isolating}, were found to have this property. This is induced by regularizing independence between latent entries or factors. We focus on these models to find morphological and color factors in cervical cell image datasets to interpret abnormality.

In this work, we explore an unsupervised method for learning features of normal cells and using negative log-likelihood to quantify abnormality. To interpret our latent space, we examine the factors of variation learned by our model and introduce a metric based on cross-entropy to cluster similar cells. We test our method against baselines of other unsupervised clustering algorithms. Our contributions are as follows:
\begin{itemize}
    \item We show that VAE models are capable of learning interpretable features from Pap smear dataset
    \item the trained models can discern abnormality by estimating a Gaussian on latent feature space using only normal samples
    \item additional image augmentation during training of generative models can improve separation between normal and abnormal samples
    \item a formulation of statistical distance based on cross-entropy with which agglomerative clustering outperforms conventional clustering methods
    \item Finally, our model is able to generalize to other datasets containing image of cervical cells; it is capable of separating normal and abnormal images by using the pre-trained encoder
\end{itemize}

\section{Related Studies}
This section reviews related studies concerning automated pap smear screening, OCC using variational autoencoders, and clustering using statistical distances.

\subsection{Automated Pap Smear Screening}
The development of a robust and accurate automated pap smear screening method was motivated by reducing human errors and examination time \cite{chitra2022recent}. This is achieved by quantitative assessments of either hand-crafted features  \cite{wang2019automatic} or the use of deep learning \cite{basak2021cervical,n2021deep,manna2021fuzzy}. Methods using hand-crafted features are interpretable in the sense that we can calculate the importance or weights of features, unlike deep learning wherein the trained model is usually a black box. A study that is most closely related to ours' is that of Özbay and Özbay \cite{ozbay2023interpretable} who proposed an interpretable deep learning framework for cervical cancer detection using the 'hash' layer. Unlike their method, by virtue of variational autoencoder formulation, we are able to determine the qualitative meaning of each feature dimension by interpreting the images generated by the decoder along a fixed direction in feature space. This allows comparison between images by calculating the change in latent space which quantitatively show how the images are different.

\subsection{OCC on VAEs}
OCC using VAEs has the benefit of learning features for the given class in an unsupervised manner. This method has been applied to detect deepfakes based on the reconstruction loss and by comparing the latent features of the input and the reconstructed image \cite{khalid2020oc}. Another study applied the same method to intrusion detection in network flow but the anomaly score set to the reconstruction probability \cite{9113298}. The advantage of our method is that additional passes to the decoder are not required as we analyze only the distribution in the latent space by using the negative likelihood in feature space as our score. To the best of our knowledge, we are the first to train our encoder using unsupervised learning and the first to apply OCC using VAEs in Pap smear anomaly detection.

\subsection{Clustering using Statistical Distances}
Clustering is an unsupervised method used to determine groups of related samples where interpoint distances are small compared with the points outside a cluster \cite{bishop2006pattern}. In the context of Pap smear classification, several clustering methods, K-means, and its fuzzy variant \cite{bezdek1973fuzzy}, have shown relatively good performance \cite{sharma2016classification, sulaiman2015improvement}. Unlike previous studies, ours is unique because we do not use fixed points to cluster our samples. Instead, we utilized non-fixed points, mapping the data into a multi-dimensional latent space as a probability distribution. Analyzing data using non-fixed points \cite{loffler2009shape} and calculating statistical distances to compare class distribution have been studied \cite{lu2021information}. However, to our knowledge, we are the first to apply clustering on non-fixed points by means of agglomerative clustering on nonfixed points \cite{mullner2011modern} using information-theory-based pseudometric. We also show that with some additional assumptions, our pseudometric becomes equal to the Euclidean distance which provides evidence for comparable but slightly better results than those obtained using K-Means with the Euclidean distance.

\section{Methods}
In this section, we define our study's problem about obtaining a score function
on the latent space of variational autoencoders (VAEs) by first reviewing VAEs in the context of disentanglement. We then define our score function for OCC and follow up using other methods for interpreting latent space
by qualitatively assessing reconstructions and using agglomerative clustering
method to obtain similar samples from input.

\subsection{VAEs and Disentanglement}
We denote $\theta_d$ as the parameters for the decoder, $\theta_e$ as the parameters for the encoder,
$q_{\theta_d}$ as the approximate posterior parameterized as a diagonal multivariate normal distribution and $p_{\theta_g}$ as the true data-generating distribution with marginal latent distribution as a standard normal distribution. Suppose we have samples $S=\{\mathbf{x_i}\}_{i=1}^N$ generated by hidden factors $\mathbf{z}$. The evidence lower bound (ELBO) of VAE \cite{kingma2013auto} is
\begin{equation}
    \mathcal{L}(\theta_d,\theta_e)=\mathcal{L}_{recon}(\theta_d, \theta_e) - \frac{1}{|S|}\sum_{\mathbf{x}\in S}\left[D_{KL}(q_{\theta_d}(\mathbf{z}|\mathbf{x})\lVert p_{\theta_g}(\mathbf{z}))\right],
\end{equation}
where $\mathcal{L}_{recon}(\theta_d, \theta_e)=\frac{1}{|S|}\sum_{\mathbf{x}\in S}E_{q_{\theta_e}(\mathbf{z}|\mathbf{x})}(p_{\theta_d}(\mathbf{x}|\mathbf{z}))$ and $D_{KL}$ is the Kullback-Leibler divergence \cite{cover1999elements}.

Burgess et al \cite{burgess2018understanding} showed that, for curated datasets, adding an extra term $\beta$ to the ELBO helps disentangle factors of variation\cite{bengio2013representation} which may be help interpret features. The ELBO for $\beta$-VAE is as follows
\begin{equation}
    \mathcal{L}_\beta(\theta_d,\theta_e;\beta)=\mathcal{L}_{recon}(\theta_d, \theta_e) - \frac{1}{|S|}\sum_{\mathbf{x}\in S}\left[ \beta D_{KL}(q_{\theta_d}(\mathbf{z}|\mathbf{x})\lVert p_{\theta_g}(\mathbf{z}))\right]
\end{equation}

Chen et al \cite{chen2018isolating} showed that total correlation \cite{watanabe1960information}, a term found by decomposing ELBO\cite{hoffman2016elbo}, is related to the disentangling property found in $\beta$-VAE. The ELBO for their model, $\beta$-TCVAE is as follows

\begin{multline}
    \mathcal{L}_{\beta-TC}(\theta_d,\theta_e;\alpha, \beta, \gamma)=\mathcal{L}_{recon}(\theta_d, \theta_e) - \alpha I_q(\mathbf{z};\mathbf{x})\\-\beta D_{KL}(q(\mathbf{z})\lVert \prod_j q(z_j) - \gamma \sum_j D_{KL}(q(z_j)\lVert p(z_j))
\end{multline}

where $I$ is the mutual information function. $\alpha=1=\gamma$ was made constant for all experiments in this study.

In this study, we explore the possibility of interpreting latent factors for cells in Pap smear images using the aforementioned variations of VAEs. By training using only normal samples, we hypothesized that color and morphological factors may be found in the latent space so that
we can interpret results when inputs shows signs of abnormality.

\subsection{Score Function}
The ELBO suggests that our approximate posterior, $q_{\hat{\theta}_e}(\mathbf{z}|\mathbf{x})$, should be a factorized standard normal distribution after training. However, our distribution of normal samples may differ from training samples owing to augmentations applied during training. Instead, we also estimate our distribution of normal samples by a multivariate normal distribution. We define our scoring function $s$ by the negative log-likelihood of a multivariate normal distribution
\begin{equation}
    s(\mathbf{z}) = -\log\mathcal{N}(\mathbf{z};\hat{\mu},\hat{V}),
\end{equation}
where $\hat{\mu}=\frac{1}{N}\sum_{\mathbf{x}}\mu(\mathbf{x};\hat\phi_\mu)$ and $\hat{V}=\frac{1}{N-1}\sum_{\mathbf{x}}(V(\mathbf{x};\hat\phi_V)+\mu(\mathbf{x};\hat\phi_\mu)\mu(\mathbf{x};\hat\phi_\mu)^T) - \hat{\mu}\hat{\mu}^T$. $\hat\phi_\mu$ and $\hat\phi_V$ are parameters in $\hat\theta_e$.

\subsection{Interpreting Latent Space}
The formulation of VAEs suggests independent image-generating factors in latent space. By traversing along the latent entry, we can infer qualitatively the meaning of each factor in latent space.
As we modeled the image distribution as an augmented distribution, we traversed along each entry centered at the mean of the nonaugmented data distribution for 10 steps starting from a value of -4 x standard deviations (with respect to the mean) to 4 x standard deviations (with respect to the mean).

Another way we can interpret latent space is to identify samples near the encoded input. The straightforward approach is to cluster samples in latent space
and identify other samples in the cluster associated with the input. However,
a characteristic of variational autoencoders may allows us to use statistical distances instead of sampling approaches to this problem.
The reparameterization trick\cite{kingma2013auto} for a Gaussian posterior, a differentiable transformation $g_{\theta_e}$, yields
\begin{equation}
    \mathbf{z}=g_{\theta_e}(\mathbf{x},\epsilon) = \mu(\mathbf{x};\theta_e)+\sigma(\mathbf{x};\theta_e)\epsilon,
\end{equation}
where $\epsilon\sim \mathcal{N}(\mathbf{0},\mathbf{1})$.
This reveals that the latent feature vector $\mathbf{z}$ is modeled by $\mu$ with an error term with variance $\sigma^2$. Therefore, instead of using contemporary clustering methods using the mean vector or a sample from this distribution (adding the error term), we can use statistical distances.
Images in the same cluster should contain similar image characteristics owing to the proximity of their location in latent space. By incorporating statistical distances, we hypothesized that our method clusters similar images more effectively compared with contemporary methods, such as K-means with Euclidean distance, by utilizing the error of representation and factoring relative information of nearby samples.

\subsection{Cross-entropy-based Referenced Statistical Distance}
We introduce a statistical distance that is related to well-known divergences such as Jenson-Shannon divergence and Jefferys divergence, and an equality to Mahalanobis distance as a special case which is shown in Appendix A. We call this Cross-Entropy based Referenced Statistical Distance (CRSD).

Given two probability distribution, $p$ and $q$, the CRSD with respect to a probability distribution $r$ is defined as
\begin{equation}
    d_{CRSD}(p,q;r) = |CE(r,q) - CE(r,p)|
\end{equation}
where $CE(p,q)$ is the cross-entropy of $q$ relative to $p$. Henceforth, for brevity purposes we will eliminate the acronym. Based on the definition, it can be inffered that the function is symmetric, $d(p,q;r)=d(q,p;r)$, non-negative, and zero if $p=q$. However, $d(p,q;r)=0$ does not imply $p=q$ and this can be easily demonstrated by Gaussian distributions with centered $r$; $p$ and $q$ have means equal to $\mu$ and $-\mu$. The triangular inequality is also satisfied and proved using simple algebra.

We also define for multiple reference probability densities: if $R$ is a set of probability densities, the CRSD with respect to $R$ is defined by

\begin{equation}
    d(p,q;R) = \sum\limits_{r\in R} |CE(r,q) - CE(r,p)|
\end{equation}

In this study, we substitute Euclidean distance for CRSD in the agglomerative clustering algorithm \cite{mullner2011modern} when using non-fixed points. We restrict $R$ to be finite and in our experiments to define the latent distributions of a given data related to training, validation, or test fold.

\subsection{Dataset}
An open conventional Pap smear dataset provided by the Center for Recognition and Inspection of Cells (CRIC) platform\cite{rezende2021cric} was chosen for our experiment. This dataset contains 400 Pap smear images and 11,534 classified cells which contains more classified cells than other open datasets \cite{martin2003pap,plissiti2018sipakmed}. The dataset contains six classes according to the Bethesda system \cite{solomon20022001}: negative for intraepithelial lesion or malignancy (NILM), low-grade epithelial lesion (LSIL), high-grade epithelial lesion (HSIL), atypical squamous cell of undetermined significance (ASC-US), atypical squamous cell cannot rule our HSIL (ASC-H), and squamous cell carcinoma (SCC). As the classified cells were annotated rather than cropped to separate images, different views of cells (achieved by rotation and displacement) could be obtained during training without introducing background elements generated by augmentation on cropped images. In addition, we have removed some annotations that were close to the boundary of the image (within 128 pixels in either x or y axis). The number of cropped images used in our experiment is shown in Table \ref{tbl:classcounts}.

\begin{table}[]
\begin{center}
\begin{tabular}{ccc}
Dataset & Class  & Counts                  \\ \hline
\multirow{6}{*}{CRIC} & NILM   & 5422                    \\
& HSIL   & 1,609                    \\
& LSIL   & 1,287                    \\
& ASC-H  & 894                     \\
& ASC-US & 563                     \\
& SCC    & 156 \\\hline
\multirow{5}{*}{SIPAKMED} & Superficial-Intermediate & 831\\
& Parabasal & 782\\
& Koliocytotic & 814\\
& Dyskeratotic & 794 \\
& Metaplastic & 785 \\\hline
\multirow{2}{*}{In-house} & Normal & 3,336,005 \\
& HSIL+ASC-H & 17,970 \\
\end{tabular}
\end{center}
\caption{Number of samples per class for each dataset used in this study.}
\label{tbl:classcounts}
\end{table}

\begin{figure}
    \centering
    \includegraphics[width=\textwidth]{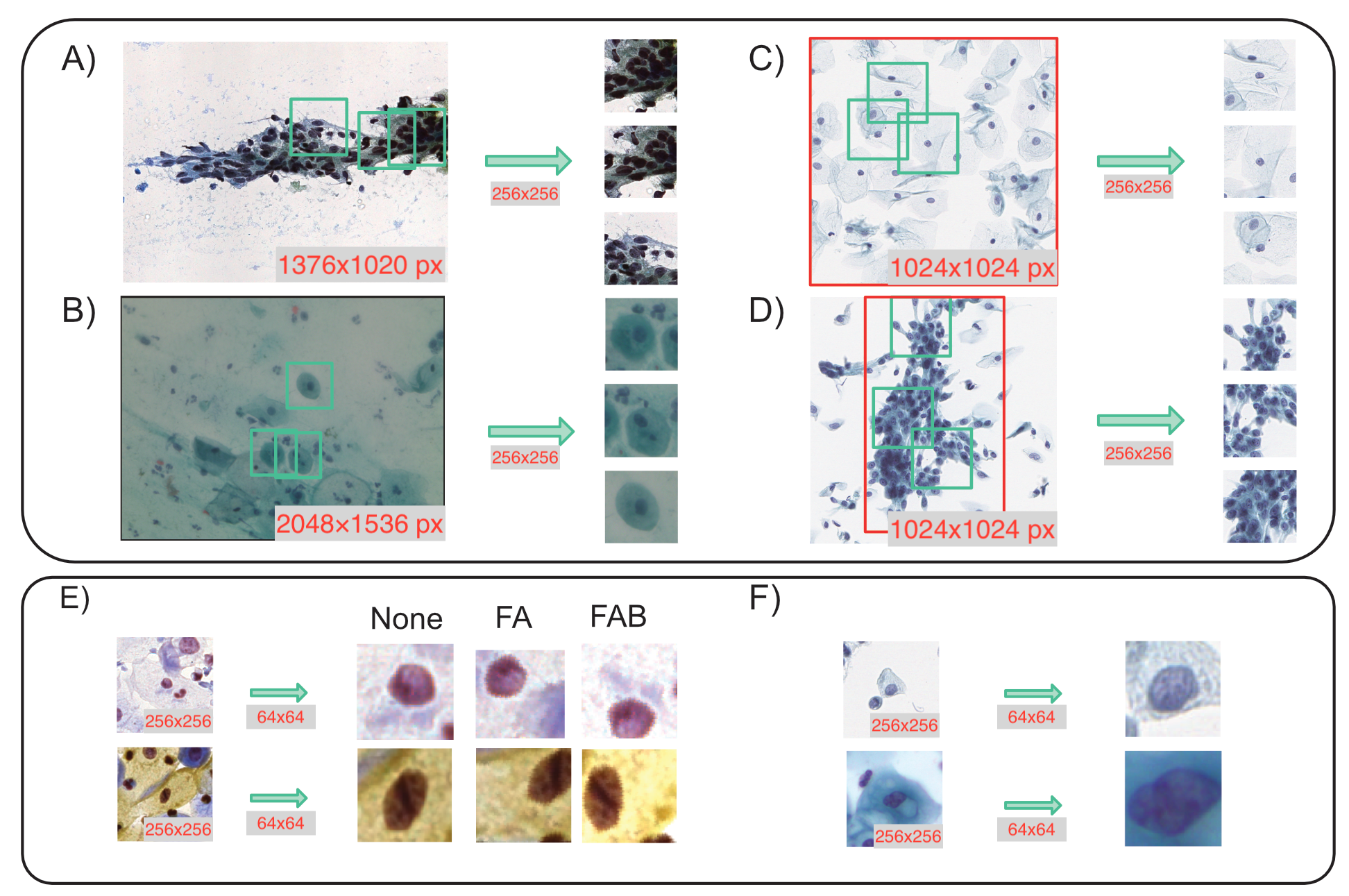}
    \caption{Example images of datasets used in our work. Tile images were cropped according to location of cells (TOP): A) CRIC Dataset, B) SIPAKMED, C) In-house normal, D) In-house abnormal. For in-house dataset, we only crop cells within the bounding box (shown in red). The cropped image were augmented further to capture the cell morphology which includes the nucleus and its immediate surroundings. E) Examples of images used for training. We applied three different augmentation techniques: None, random horizontal-flip and affine (FA), random horizontal-flip, affine, and bightness adjustment (FAB). F) Examples of in-house (top) and SIPAKMED (bottom) images used for external evaluation.}
    \label{fig:overview2}
\end{figure}

To check for the generalizability of our model, we used SIPAKMED \cite{plissiti2018sipakmed} and an in-house dataset to compare scores between abnormal and normal cells. The SIPAKMED dataset contained information on cell morphological information which we used to determine the cell coordinates. We have retrospectively collected 100 normal and 50 abnormal patients including HSIL and ASC-H cells under the approval Daegu Joint Institutional Review Board (DGIRB-2021-11-002). Our data consisted of whole-slide images of liquid-based pap smears without precise bounding boxes for individual cells. Whole-slide images containing abnormal regions were annotated. To obtain bounding boxes for individual cells, we used the Segment Anything Model (SAM) to detect cell locations \cite{kirillov2023segment}. From the 100 normal whole slide images and 26 abnormal whole slide images, we applied SAM to 1024 by 1024 tiled images to obtain our cell images. To obtain only abnormal cells from abnormal whole-slide images, we detected cells only within manually annotated regions containing abnormal cells. The dimension of all cell images were is 256 by 256; these images were center cropped to 64x64 before they were input into the model. This limits the field of view of our model to focus on the nucleus of a single cell and its immediate surroundings. Figure \ref{fig:overview2} shows the cropped images which were used as input.

\subsection{Experiments}
\begin{figure}
    \centering
    \includegraphics[width=\textwidth]{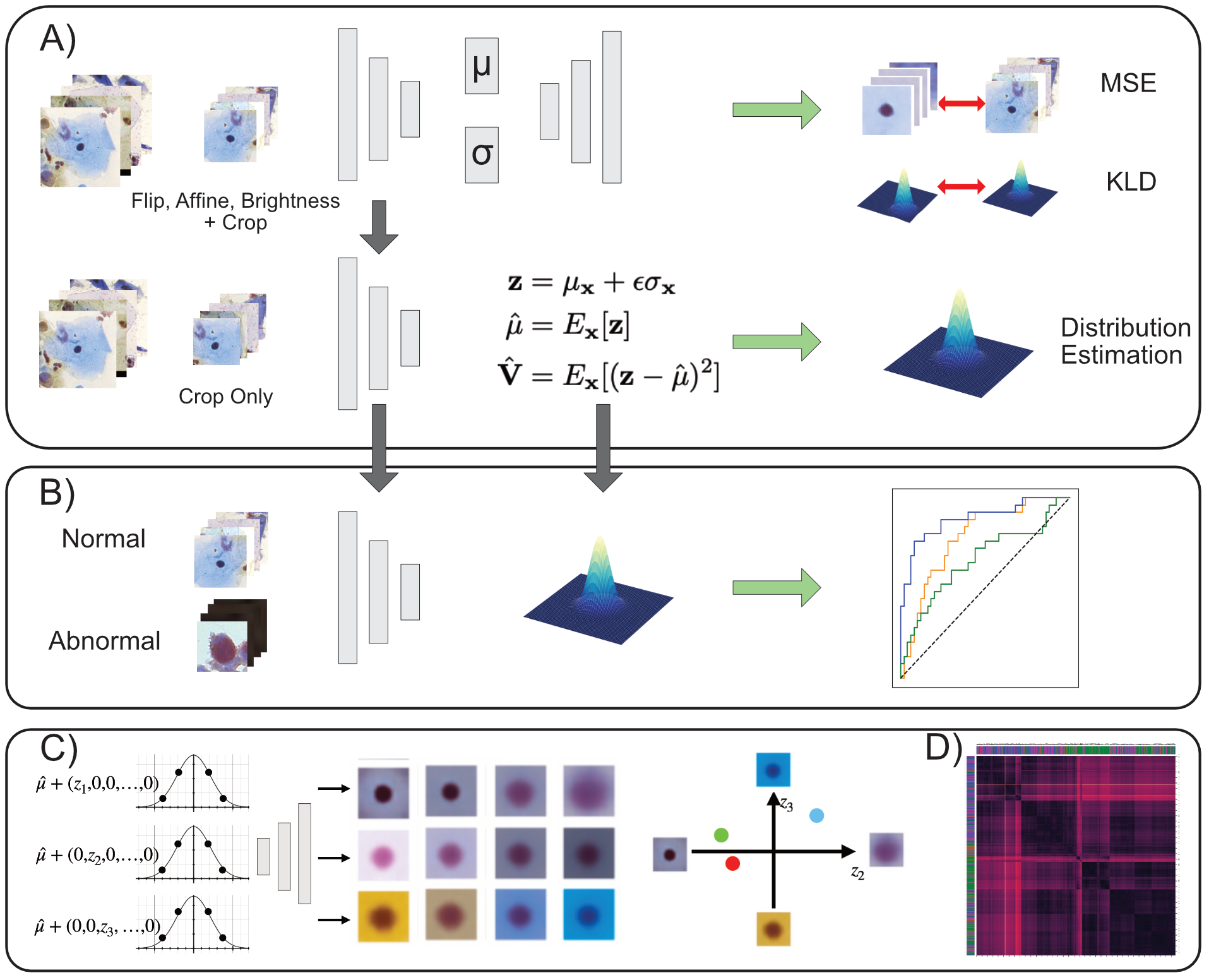}    
    \caption{Overview of Interpretable Cervical Cancer Screening using VAE. (a) Variants of VAE are trained on the CRIC dataset. After training, we estimate parameters of multivariate guassian distribution. (b) To screen for abnormality, we use negative log-likelihood in latent space using our estimated distribution obtained in (a). (c) Generative latent variables are characterized by their generating factor assesed qualitatively after training the model. Images were generated by traversing in orthogonal directions with equal steps. (d) Agglomerative clustering is performed
    with a statistical pseudo-distance to cluster samples containing both normal and abnormal samples. Similar samples, with
    respect to latent space, will be close to each other.}
    \label{fig:overview}
\end{figure}

Our study consists largely of four parts: training of VAEs and the estimation of normal sample distribution in latent space, abnormality scoring, interpretation method for latent representations, and clustering. An overview is shown in Figure \ref{fig:overview}.

We trained VAE, $\beta$-VAE, and $\beta$-TCVAE using normal samples (NILM) in the CRIC
dataset to learn disentangled features and to calculate a simple scoring function to discern abnormal samples. 
Five-fold cross-validation was performed on a two class classification task (NILM vs. rest) to identify the best hyperparameters. The hyperparameters were varied using $\beta=1,4,16$, and latent dimensions 8, 32, and 128.

Additionally, image augmentation techniques were varied to determine if additional image preprocessing had a positive effect on model performance. In generative models, severe augmentations tend to distort the learned data distribution; thus we selected plausible augmentations, such as rotation, translation, and brightness adjustment, that mimic possible variations in the usual data acquisition process using an optical microscope. Three methods were tested as hyperparameters in the cross-validation process: none, random flip, rotation and translation (FA), and random flip, rotation, translation and brightness (FAB). Example images are shown in Figure \ref{fig:overview2}. For each abnormal class, scores were then compared against NILM to assess the per-class performance of our model. Finally, 
clustering algorithms were compared using our best model. Clustering algorithms included agglomerative clustering \cite{mullner2011modern}, K-means \cite{ahmed2020k}, spectral clustering\cite{ng2001spectral}, and DBSCAN \cite{ester1996density}. We chose homogeneity, completeness and $V$-measure\cite{rosenberg2007v} as a metric for comparison. We conducted tests using 100 clusters and other parameters were set to default using sci-kit learn implementation \cite{sklearn_api}.

We experimented on additional external datasets to see if our model could score abnormality with comparable or better performance. The model was fixed, only to encode images to latent space. The distribution of normal images was estimated using the same procedure as with the CRIC dataset. The datasets used in the futher analysis were the SIPAKMED dataset and in house dataset obtained from Kyungpook National University Chilgok Hospital. Unlike the SIPAKMED dataset, the in-house dataset consisted of whole slide images which required additional preprocessing in order to obtain cropped images of individual cells. Bounding boxes containing HSIL were also given. We used Segment Anything model (SAM)\cite{kirillov2023segment} to generate masks of cells. The center of the contour was used as a reference point to crop a 256x256 image. For abnormal cells, we applied a mask generated by bounding boxes from which candidate abnormal cells were obtained using SAM. In total, we found 3,336,005 normal cell crops and 17,970 abnormal cell crops. During our evaluation process, we used repeated random sampling of normal cells and abnormal cells. 6,000 normal cells were sampled of which 3,000 was used for estimating the Gaussian and the abnormality score was calculated on the other 3,000 normal cells and 3,000 abnormal cells. For SIPAKMED, we estimated Gaussian on different folds of normal data (5-fold split on training datasets). Abnormal data were split into 5-folds each evaluated against normal data to calculate the receiver operating characteristic (ROC) curve.  

\section{Results}

\subsection{Hyperparameter Tuning}
\begin{figure}
    \centering
    \includegraphics[width=0.9\textwidth]{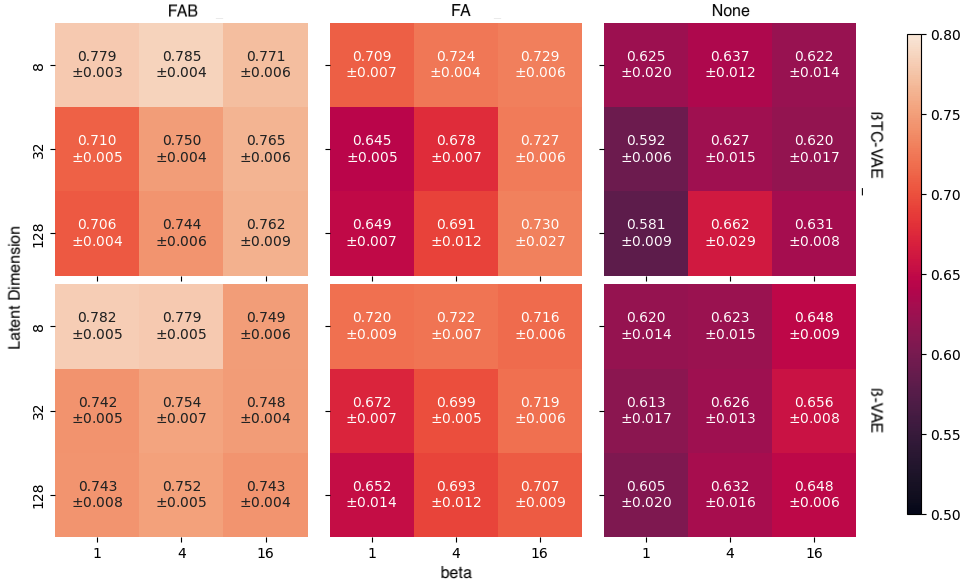}
    \caption{Cross-fold AUROC) ($\pm$ standard deviation.) for each hyperparameter (latent dimension, $\beta$), data augmentation (None, FA or FAB), and model configuration. Applying extra augmentation (in particular the combination of horizontal, rotation, and brightness augmentations), AUROC increased in all hyperparameteric combination cases. The best model was TCVAE with $\beta$=4 and a latent dimension of eight.}
    \label{fig:hyperparameter}
\end{figure}

In supervised learning, additional augmentations tend to enhance the performances of deep models but this may not directly translate to
generative models because augmented samples distort the target data distribution to be learned. Therefore, our experiments were conducted on relatively plausible augmentations: rotation of cells, systematic error in centering a cell in an image, and brightness configuration which depends on the light source of the microscope. Figure \ref{fig:hyperparameter} shows that for any hyperparameter combination, using the full augmentation (FAB) yields the best area under the receiver operating characteristic curve (AUROC) on average. The best model was found to be $\beta$-TCVAE with $\beta=4$ and latent dimension of 8. Further analysis was performed using this model.

\subsection{Latent Feature Space}
\begin{figure}
    \centering
    \includegraphics[width=0.6\textwidth]{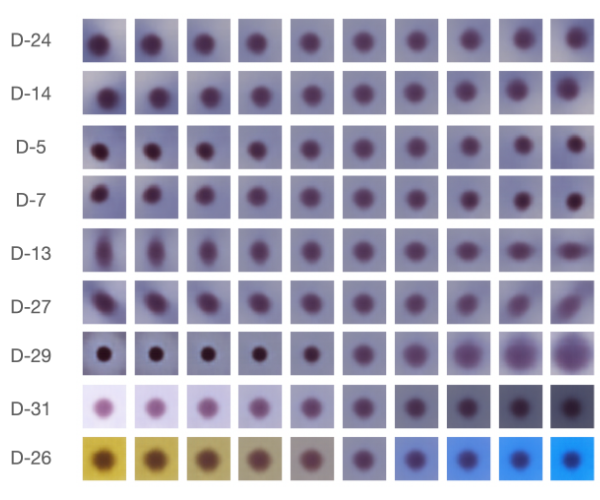}
    \caption{Unsupervised Morphological and Color Features. Features learned by a model, $\beta$-VAE with $\beta=4$ and latent dimension of 32 trained with FAB augmentation, are shown. Color (D-26), brightness (D-31), and several morphological features were discovered by the model.}
    \label{fig:traversal}
\end{figure}

Morphological and color-related features were learnt by the model using unsupervised training. Figure \ref{fig:traversal} shows features learnt by one of the models. The reconstructed images were obtained by calculating the mean of latent features of normal samples and traversing from -4 $\times$ standard deviations to 4 $\times$ standard deviations from the mean in 10 steps. The top seven rows correspond to morphology of the cell image concerning the locations, shapes and sizes of the nucleus. The bottom two rows correspond to color. D-31 is a brightness adjustment and D-26 is a hue adjustment from gold-brownish to blue.

\begin{figure}
    \centering
    \includegraphics[width=0.8\textwidth]{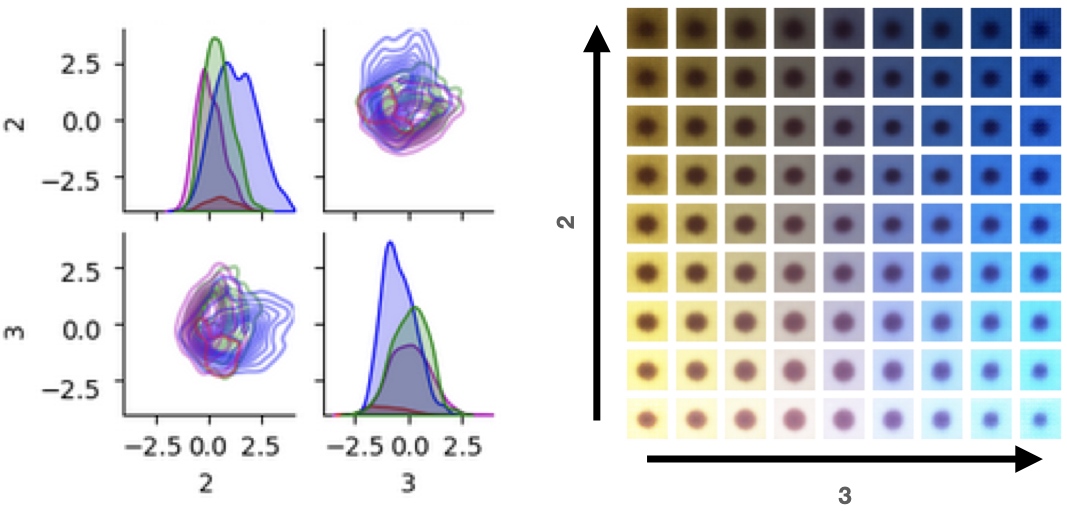}
    \caption{Pairwise KDE for latent dimensions 2 and 3 with generated images in this plane. KDE plots (left) show that different classes have different distributions. The classes are NILM (magenta), LSIL+ASC-US (green), HSIL+ASC-H (blue), and SCC (red). With reference to the decoded images, we can compare distributions of different classes. The images on the right are decoded images in the latent plane (latent features 2 and 3). The rows correspond to latent dimension 3 with increasing value from left to right and the columns correspond to latent dimension 2 increasing in value from bottom to top. KDE plots for the other dimensions are in Supplementary Figure A1.
    }
    \label{fig:pairwisekde}
\end{figure}

Pairwise KDE plots are shown for selected two latent features in Figure \ref{fig:pairwisekde}. Each abnormal class contains different image characteristics which is evident by differences in densities between classes. Notice that our estimated normal sample distribution is not a standard Gaussian. KDE plots for the other dimensions are in Supplementary Figure A1.

\subsection{Class-wise Screening Detection}
We also calculated the performances of our scoring function by comparing each abnormal class against normal samples. Table \ref{tbl:classscores} lists accuracy, AUC, F1 scores, sensitivity and specificity outcomes for each abnormal class. The model performed well at discerning SCC and HSIL from normals followed by ASC-H, LSIL, and ASC-US. The AUC score corresponds to severity of the lesion; AUC scores are ordered from high grades HSIL, SCC, and ASC-H, to low grades LSIL and ASC-US.
\begin{table}[]
\centering
\resizebox{0.8\textwidth}{!}{
\begin{tabular}{cccccc} class
       & Accuracy                                               & AUC                                                    & F1 score                                               & Sensitivity                                            & Specificity                                            \\ \hline
SCC    & \begin{tabular}[c]{@{}c@{}}0.806±0.029\end{tabular} & \begin{tabular}[c]{@{}c@{}}0.908±0.003\end{tabular} & \begin{tabular}[c]{@{}c@{}}0.883±0.020\end{tabular} & \begin{tabular}[c]{@{}c@{}}0.800±0.035\end{tabular} & \begin{tabular}[c]{@{}c@{}}0.870±0.036\end{tabular} \\
HSIL   & \begin{tabular}[c]{@{}c@{}}0.849±0.002\end{tabular} & \begin{tabular}[c]{@{}c@{}}0.920±0.002\end{tabular} & \begin{tabular}[c]{@{}c@{}}0.850±0.006\end{tabular} & \begin{tabular}[c]{@{}c@{}}0.829±0.027\end{tabular} & \begin{tabular}[c]{@{}c@{}}0.871±0.026\end{tabular} \\
ASC-H  & \begin{tabular}[c]{@{}c@{}}0.742±0.005\end{tabular} & \begin{tabular}[c]{@{}c@{}}0.808±0.006\end{tabular} & \begin{tabular}[c]{@{}c@{}}0.787±0.007\end{tabular} & \begin{tabular}[c]{@{}c@{}}0.731±0.018\end{tabular} & \begin{tabular}[c]{@{}c@{}}0.761±0.020\end{tabular} \\
LSIL   & \begin{tabular}[c]{@{}c@{}}0.647±0.006\end{tabular} & \begin{tabular}[c]{@{}c@{}}0.689±0.005\end{tabular} & \begin{tabular}[c]{@{}c@{}}0.684±0.018\end{tabular} & \begin{tabular}[c]{@{}c@{}}0.672±0.044\end{tabular} & \begin{tabular}[c]{@{}c@{}}0.613±0.043\end{tabular} \\
ASC-US & \begin{tabular}[c]{@{}c@{}}0.449±0.030\end{tabular} & \begin{tabular}[c]{@{}c@{}}0.549±0.013\end{tabular} & \begin{tabular}[c]{@{}c@{}}0.492±0.055\end{tabular} & \begin{tabular}[c]{@{}c@{}}0.358±0.058\end{tabular} & \begin{tabular}[c]{@{}c@{}}0.724±0.055\end{tabular}
\end{tabular}
}
\caption{Classwise Metrics for Best Model. We calculated metrics for each abnormal class compared to normal dataset. The model was good at discerning SCC and HSIL from normals followed by ASC-H, LSIL, and ASC-US.}
\label{tbl:classscores}
\end{table}

\subsection{Clustering Performance}

Cross-validation (5-fold) on the abnormal dataset was performed to
compare different unsupervised algorithms using best model. Table \ref{tbl:algorithm} shows that agglomerative methods yield higher $V$ measures than other methods with the agglomerative clustering with statistical metric (Agg-SM) as the best performing algorithm. Homogeniety for aggomerative clustering with Euclidean distance (Agg-EM) and KMeans were similar but KMeans had lower $V$ measure. Spectral clustering and DBSCAN had very low $V$ measures compared to other algorithms but DBSCAN had highest completeness score suggesting that the algorithm favors smaller number of clusters.

\subsection{External Validation}
\begin{table}[]
\resizebox{\textwidth}{!}{
\begin{tabular}{ccccc}
\textbf{Algorithm} & \textbf{h}            & \textbf{c}        & \textbf{V}        \\ \hline
Agg-SM          & 0.393$\pm$0.0117       & 0.151$\pm$0.00259  & 0.218$\pm$0.005                     \\                  
Agg-EM & 0.282$\pm$0.00591 & 0.0940$\pm$0.00174 &  0.141 $\pm$ 0.00265 &   \\
KMeans     & 0.275$\pm$0.006       & 0.0725$\pm$0.0019 & 0.114$\pm$0.003                                  \\
Spectral Clustering       & 0.0526$\pm$0.0031     & 0.0950$\pm$0.0044 & 0.0676$\pm$0.0025               \\
DBSCAN   & 0.000234$\pm$0.000413 & 0.746$\pm$0.414   &0.000463$\pm$0.000818             
\end{tabular}%
}
\caption{Unsupervised Clustering Algorithm Comparisons. Homogeneity (h), completeness (c) and V-measure (V) are calculated for each algorithm with number of clusters used. Our method, Agg-SM, gives best V-measure followed by K-Means. Homogeneity was similar between agglomerative clustering using Euclidean distance (Agg-EM) and K-Means.}
\label{tbl:algorithm}
\end{table}

\begin{figure}
    \centering
    \includegraphics[width=\textwidth]{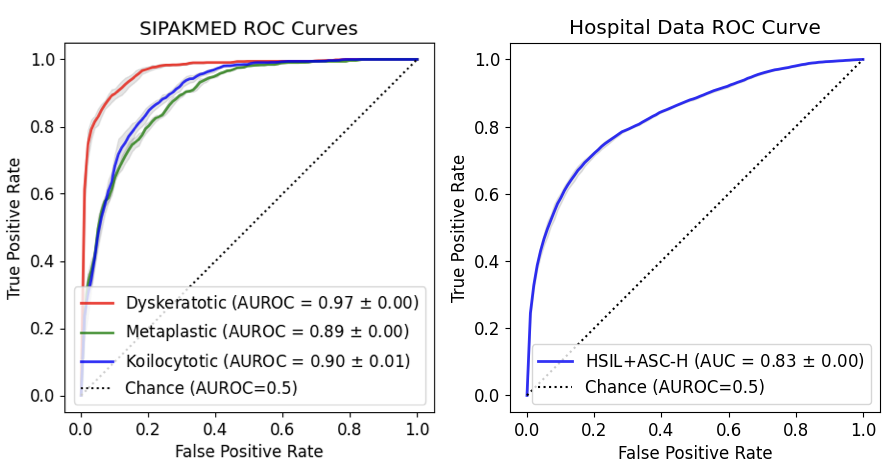}
    \caption{ROC curves using SIPAKMED and in-house datasets}
    \label{fig:AUROC}
\end{figure}

Finally, we evaluated the AUROC of discerning abnormality from normals by using the score function for additional datasets: SIPAKMED and the in-house dataset. Figure \ref{fig:AUROC} shows receiver operating characteristic curves of each abnormal class for each dataset. On average, the AUC values for dyskeratotic, metaplastic, and koliocytotic cells were 0.97, 0.89, and 0.9 respectively. The AUC value for HSIL+ASC-H on in-house data was 0.83. Our method shows good generalizing performance outcomes when used on external datasets.

\section{Discussion}
VAEs enable the disentanglement of latent features that allow modeling of the latent space of cervical cells in Pap smear images. Experiments showed that we were successful in obtaining features corresponding to cytoplasmic color, nuclear shape, and image brightness in an unsupervised manner while obtaining respectable detection for SCC and HSIL. More importantly, a simple Gaussian estimation on the latent feature space of normal examples produced these results without the requirement of additional highly parameterized model training.

Adding additional augmentations during the training of a generative model improved the discerning capability. However, as the generative model learns the data distribution, applying image augmentation can distort the original data distribution \cite{jun2020distribution}. This is also shown in Supplementary Figure A1, where the normal class (NILM) is not represented by a centerted gaussian distribution for each latent entry. We taken account of this change in latent distribution by fitting a Gaussian distribution on the normal samples rather than using the assumed prior of the data distribution.

A comparison of our results with those of other studies showed that there is a trade-off using limited number of labeled samples for deep feature learning. Table \ref{tbl:twoclass} shows the performance outcomes of other models with supervised training. Despite training with only normal examples, our model was able to discern abnormality with 0.7 accuracy in binary classification task. This may be due to low AUC scores for low grade lesions (LSIL and ASC-US) as seen from Table \ref{tbl:classscores}, which contributes to overall lower performance.
\begin{table}[]
\resizebox{\textwidth}{!}{
\begin{tabular}{llllll}
                                      & Sensitivity & Specificity & Accuracy & F1 & Training Method \\ \hline
Diniz et al. (Random Forest)\cite{diniz2021hierarchical} & 0.958                & 0.958                & 0.958             & 0.958       & Supervised                                    \\
Diniz et al. (Ensemble)\cite{n2021deep}    & 0.96                 & 0.96                 & 0.96              & 0.96        & Supervised                                       \\
CVM-Cervix \cite{liu2022cvm}            & 0.916                &                      & 0.9172            & 0.852       & Supervised                              \\
Ours ($\beta-tcvae$-b4d8)   & 0.759                & 0.683                & 0.70              & 0.58        & OCC                                      
\end{tabular}
}
\label{tbl:twoclass}
\caption{Binary classification results. OCC with VAEs perform worse than supervised training.}
\end{table}

This study was associated with some limitations. While our model can detect abnormalities for most classes, it does not discriminate between abnormal classes which supervised models do effectively. To overcome this, we devised a querying method using agglomerative clustering. Images contained in some single cluster will have similar features. Although we cannot label a particular cluster as some particular abnormality, related images can be queried and morphological and image related features can be assessed quantitatively to assist pathologists examining cells in Pap smears. Despite the fact that we have shown our model was effective at abnormality detection on different datasets to the training dataset, external data will contain different latent distributions depending on acquisition method as seen by comparing latent spaces for CRIC, in-house and SIPAKMED ( Supplementary Figure A1, A2, 
and A3
respectively).

\section{Conclusion}
VAEs provide means to interpret latent features by
examining the reconstructed output and give a simple formulation for scoring abnormality. OCC of normal
cells in Pap smear images can predict cell abnormality without using abnormal labels. Performance was shown to increase when applying additional image
augmentation during training. Along with interpreting latent features directly,
a cross-entropy based metric was introduced for agglomerative clustering. The
method consistently outperformed other common unsupervised clustering algorithms. We have shown that, although binary classification performance is lower than supervised methods, using
OCC can discern HSIL and SCC from normals with high accuracy. We also demonstated a method for explainability that may contribute to the development of explainable artificial inteligence (AI) and AI trust in healthcare.

\section{Acknowledgements}
This work was supported by the Brain Pool program funded by the Ministry of Science and ICT through the National Research Foundation of Korea [2020H1D3A2A02102040, 2022H1D3A2A01096490]; Ministry of Education [2021R1I1A3056903]

\appendix
\section{Relation of CSRD to other divergences}

The distance is equal to the Kullback-Leibler, Jefferys, and Jenson-Shannon divergences depending on which reference probability distribution or a set of probabilities is chosen. We denote the divergences as $D_{KL}$, $D_J$, and $D_{JSD}$, respectively. The following are some relations of our formulation to well-known divergences \cite{61115}
\begin{align}
    d(p,q;p) &= D_{KL}(p\lVert q) \\
    d(p,q;\{p,q\}) &= 2 D_{J}(p,q) \\
    d(p,q; \{p,q\}) &\ge 4 D_{JSD}(p,q)
\end{align}
If $p$, $q$, are Gaussian distributions with equal covariances, then we express the squared Mahalonobis distance as
\begin{equation}
    d(p,q;\{p,q\}) = (\mu_p-\mu_q)^T\Sigma^{-1}(\mu_p-\mu_q).
\end{equation}
Additionally, if $\Sigma=I$, we are then essentially calculating distances between latent distributions by the Euclidean norm between the means, i.e. $d(p,q;\{p,q\})=\lVert\mu_p-\mu_q\rVert_2^2$.
Therefore, a connection is made to our method using standard clustering algorithms using the mean vectors of the latent representations to our method of using statistical distances; the former is a special case of the latter with assumptions on the covariances and the choice of the reference distributions.

 \bibliographystyle{elsarticle-num} 
 \bibliography{main}





\end{document}